\pdfoutput=1
\documentclass[a4paper, 10 pt, conference]{ieeeconf}  
\IEEEoverridecommandlockouts                              

\usepackage{graphicx}
\usepackage[hidelinks]{hyperref}

\usepackage{graphics} 
\usepackage{graphicx}
\usepackage[ruled,vlined,linesnumbered,noresetcount]{algorithm2e}
\usepackage[]{units}
\usepackage{pgfplots}
\usepackage{tikz}
\usetikzlibrary{positioning}
\usetikzlibrary{shapes.geometric}
\usetikzlibrary{shapes.misc}
\usetikzlibrary{external}
\usepackage{paralist}
\usepackage{mathtools}

\usepackage{xcolor}
\usepackage{siunitx}
\usepackage{multirow}
\usepackage{epstopdf}
\graphicspath{ {Pictures/} }
\usepackage{epstopdf}
\usepackage{tikz}
\usepackage{pgfplots}
\usetikzlibrary{shapes,arrows,patterns}
\usetikzlibrary{arrows.meta}
\usetikzlibrary{plotmarks}
\usetikzlibrary{calc,decorations.markings,math}
\usetikzlibrary{external}
\pgfplotsset{compat=1.14}
\usepackage{grffile}

  \usepgfplotslibrary{patchplots}
\usepackage{epsfig}
\usepackage{float}

\usepackage{amsmath} 
\usepackage{amssymb}  
\usepackage{mathtools}
\graphicspath{ {Pictures/} }
\DeclareGraphicsExtensions{.pdf,.eps}

\usepackage[acronym,nomain,nonumberlist]{glossaries}

\newacronym{ahe}{AHE}{Adaptive Histogram Equalization}
\newacronym{ar}{AR}{Augmented Reality}
\newacronym{clahe}{CLAHE}{Contrast Limited Adaptive Histogram Equalization}
\newacronym{cnn}{CNN}{Convolutional Neural Network}
\newacronym{dbscan}{DBSCAN}{Density-Based Spatial Clustering of Applications with Noise }
\newacronym{dl}{DL}{Deep Learning}
\newacronym{dof}{DoF}{Degree of Freedom}
\newacronym{ekf}{EKF}{Extended Kalman Filter}
\newacronym{fbe}{FBE}{Forward Backward Envelope}
\newacronym{fov}{FoV}{Field of View}
\newacronym{fps}{fps}{Frame Per Second}
\newacronym{gnss}{GNSS}{Global Navigation Satellite System}
\newacronym{gps}{GPS}{Global Positioning System}
\newacronym{iid}{IID}{Independent Identically Distributed}
\newacronym{imu}{IMU}{Inertial Measurement Unit}
\newacronym{kf}{KF}{Extended Kalman Filter}
\newacronym{llwcf}{LLWCF}{Left Local Wall Continuum Function}
\newacronym{lrcn}{LRCN}{ Long-Term Recurrent Convolutional Network}
\newacronym{lwcf}{LWCF}{Local Wall Continuum Function}
\newacronym{mae}{MAE}{mean absolute error}
\newacronym{mav}{MAV}{Micro Aerial Vehicle}
\newacronym{mhe}{MHE}{Moving Horizon Estimation}
\newacronym{mimo}{MIMO}{Multiple Input Multiple Output}
\newacronym{mlp}{MLP}{MultiLayer Perceptron}
\newacronym{mpc}{MPC}{Model Predictive Control}
\newacronym{msf}{MSF}{Multi Sensor Fusion}
\newacronym{nmhe}{NMHE}{Nonlinear Moving Horizon Estimation}
\newacronym{nmpc}{NMPC}{Nonlinear Model Predictive Control}
\newacronym{nn}{NN}{Nueral Network}
\newacronym{nwu}{NWU}{North-West-Up}
\newacronym{panoc}{PANOC}{Proximal Averaged Newton-type method for Optimal Control}
\newacronym{pdf}{PDF}{Probability Density Function}
\newacronym{pid}{PID}{Proportional Integral Derivative}
\newacronym{psd}{PSD}{ Positive Semi-Definite }
\newacronym{rbf}{RBF}{ Radial Basis Function}
\newacronym{relu}{ReLU}{Rectified Linear Unit}
\newacronym{rl}{RL}{Reinforcement Learning}
\newacronym{rlwcf}{RLWCF}{Right Local Wall Continuum Function}
\newacronym{rmse}{RMSE}{Root Mean Square Error}
\newacronym{ros}{ROS}{Robot Operating System}
\newacronym{sfm}{SfM}{Structure from Motion}
\newacronym{sqp}{SQP}{Sequential Quadratic Programming}
\newacronym{ugv}{UGV}{Unmanned Ground Vehicle}
\newacronym{ukf}{UKF}{Unscented Kalman Filter}
\newacronym{vi}{VI}{Visual Inertia}
\newacronym{vo}{VO}{Visual Odometry}

\usepackage{url}
\newlength\fwidth

\title{ \LARGE \bf Unsupervised Learning for Subterranean Junction Recognition Based on 2D Point Cloud\thanks{This work has been partially funded by the European Unions Horizon 2020 Research and Innovation Programme under the Grant Agreement No. 730302 SIMS. Funding from Vinnova in  the project ‘AI Factory for Railway’ is also acknowledged.  Corresponding author's e-mail: sinsha@ltu.se}}
\author{Sina Sharif Mansouri$^{1}$, Farhad Pourkamali-Anaraki$^{2}$, Miguel Castaño Arranz$^{3}$, Ali-akbar Agha-mohammadi$^{4}$, \\ Joel Burdick$^{5}$, and George Nikolakopoulos$^{1}$ 
\thanks{$^{1}$ Robotics and AI Team, Department of Computer, Electrical and Space Engineering, Lule\r{a} University of Technology, Lule\r{a} SE-97187, Sweden}
\thanks{$^{2}$ Department of Computer Science, University of Massachusetts Lowell, MA, USA}%
\thanks{$^{3}$ Department of Civil, Environmental and Natural Resources Engineering Lule\r{a} University of Technology, Lule\r{a} SE-97187, Sweden.}%
\thanks{$^{4}$ The author is with Jet Propulsion Laboratory California Institute of
Technology Pasadena, CA, 91109.}%
\thanks{$^{5}$ Division of Engineering and Applied Sciences, California Institute of Technology, Pasadena, California, USA.}%
}
\SetKwInOut{Parameter}{Parameter}

\begin{document}
\maketitle       
\maketitle
\thispagestyle{empty}
\pagestyle{empty}

\begin{abstract}
This article proposes a novel unsupervised learning framework for detecting the number of tunnel junctions in subterranean environments based on acquired 2D point clouds. The implementation of the framework provides valuable information for high level mission planners to navigate an aerial platform in unknown areas or robot homing missions. The  framework utilizes spectral clustering, which is capable of uncovering hidden structures from connected data points lying on non-linear manifolds. The spectral clustering algorithm computes a spectral embedding of the original 2D point cloud by utilizing the eigen decomposition of a matrix that is derived from the pairwise similarities of these points. We validate the developed framework using multiple data-sets, collected from multiple realistic simulations, as well as from real flights in underground environments, demonstrating the performance and merits of the proposed methodology.  
\end{abstract}


\glsresetall
%
%
%
\section{Introduction}
Recent technological advances in the field of robotics lead to the deployment of \glspl{mav} in challenging environments such as underground mine navigation~\cite{mansouri2019visioncnn}, search and rescue mission~\cite{tomic2012toward}, large infrastructure inspections~\cite{mansouri2018cooperative}, to name a few. \glspl{mav} have the potential to navigate rapidly in complex, unpredictable, and diverse subterranean environments for collecting various data types or saving lives.

\begin{figure*}[htbp!]
 \centering
  \includegraphics[width=\linewidth]{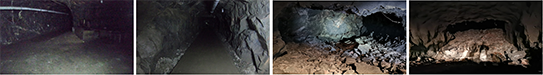} 
   \caption{Photos of an underground mine in Sweden illustrating complex geometry, irregular geological structure, etc.}
    \label{fig:conceptmine}    
\end{figure*}

Despite the broad spectrum of technological advances in the field, autonomous navigation in subterranean environments is still an ongoing quest. Subterranean environments face unique challenges, including irregular geological structures, complex geometry, lack of \gls{gps} signal, the general absence of illumination, constrained passages, and unpredictable topology (see Figure~\ref{fig:conceptmine} for illustration). Complex geometry and the presence of multiple junctions hinder the navigation mission. Junction detection is a critical task to enable safer and more optimal overall mission execution. For this task, a 2D Lidar equipped on the \glspl{mav} is used for sensing the environment. The high-end 3D lidars can provide 3D point cloud, and it can be used for junction detection. However, we are trying to remove the dependency on high end and highly expensive sensors, replacing them with lower cost solutions. Visual sensors cannot provide sufficient information regarding the complex geometry of subterranean environments due to lack of illumination. Furthermore, burdening a \gls{mav} with optical sensors in all directions and a bright light source will reduce the flight time, and require a large computational resource to post-process the visual feedback. 

The 2D point cloud extracted from a 2D lidar is used in this paper to obtain important information about the number of junctions. The introduced method relies on a family of unsupervised learning techniques that compute a spectral embedding of the original data points by forming similarity graphs. To be more specific, spectral clustering algorithms typically consist of three steps~\cite{von2007tutorial}. The first step involves forming a similarity graph that encodes information about all pairwise similarities among data points in the given 2D point cloud. A popular approach for defining such similarities is to employ the \gls{rbf}~\cite{pourkamali2019improved}. The next step is to compute the eigenvalue decomposition of the normalized Laplacian matrix that will provide a non-linear mapping of the original point cloud~\cite{tremblay2020approximating}. A key feature of our proposed framework is that the number of zero eigenvalues of the Laplacian reveals the number of connected components of the constructed similarity graph. The third step is to use the \textit{K-means} clustering algorithm~\cite{pourkamali2017preconditioned} to partition the 2D point cloud. Hence, our proposed method partitions the transformed 2D point cloud into connected components or clusters for identifying the number of junctions. 

\subsection{Background \& Motivation}
Navigation based on visual sensors or 2D/3D lidar has received significant attention in recent years in different application scenarios, e.g.,~\cite{Kanellakis2017,barfoot2016into}. One of the essential capabilities for successful autonomous navigation in unknown and uncertain environments is environmental awareness, such as identifying environment geometries, obstacles, junctions, dead-ends, etc. 

Several works have considered the use of visual sensors, lidar, and satellite images for detecting road intersections in uncertain environments~\cite{chaudhuri2012semi}. For example, the previous work \cite{chen2019higher} developed an automatic road junction detection technique using airborne lidar data as well as the direction and width of road branches. Synthetic aperture radar systems are used for detecting L- and T- shape road junctions in~\cite{negri2006junction} based on Markov optimization. The authors of~\cite{habermann2016road} developed a machine learning technique based on the 3D point cloud generated by a laser rangefinder to detect road junctions. The introduced method was evaluated on two different data-sets from Germany and Brazil roads. The studied frames were classified into two groups of ``junctions'' or ``roads". In~\cite{mueller2011gis}, junction detection is used to improve map localization in areas with weak or erroneous \gls{gps} signals such as forests or urban canyon. The proposed method detected crossroads using the lidar data by identifying free spaces between obstacles. However, most of these methods cannot be used in mines, as the map of such environments is not readily available, and existing methods are only suitable for certain shapes of junctions. Therefore, the prior work on using the lidar data for identifying subterranean junctions has demonstrated only limited success.

Towards using visual feedback for junction detection, in \cite{kumaar2018juncnet}, a \gls{cnn} binary classifier was developed for outdoor road junction detection. The method is experimentally evaluated on a commercially available \gls{mav} Bebop 2 from Parrot. In~\cite{bhatt2017have}, a road intersection detection module has been proposed as binary classification, using the \gls{lrcn} architecture to identify relative changes in outdoor features and eventually detecting intersections. In \cite{kumar2018towards}, the authors proposed an architecture that combines \gls{cnn}, Bidirectional LSTM~\cite{bhatt2017have} and Siamese~\cite{yih2011learning} style distance function learning for junction recognition in videos. In this work, the authors evaluated their approach on different data-sets spanning various experimental scenarios. Towards junction detection in underground mines, in~\cite{mansouri2019visualjunction}, the authors proposed transfer learning with AlexNet~\cite{alexnet}. The method is limited to identifying only three different junctions and relied on the looking forward camera heavily. Moreover, these methods mostly consider binary classifiers, while in real-life scenarios more complex types of junctions exist and the junction recognition should recognize the different types of junctions. Additionally, most of these works have been evaluated and tuned in out-door environments and with proper illumination, i.e., using rich data about the surrounding of the platforms. 

In this work, we propose to identify the number of junctions in uncertain subterranean environments using an unsupervised learning framework. Cluster analysis is one of the most fundamental problems in machine learning for finding groups of similar data points without any supervision~\cite{chen2017revisiting}. Many existing clustering methods such as \textit{K-means} clustering learn hidden structures from data points that are connected within convex boundaries. However, in this work, our goal is to develop a framework for identifying junctions without posing restrictive assumptions. Thus, we propose to use a family of unsupervised learning methods known as spectral clustering~\cite{ng2002spectral}. The superior performance of spectral clustering stems from the ability to exploit non-linear pairwise similarities between data points. As a result, spectral clustering has found applications in many domains, including image segmentation and genomic data analysis~\cite{tung2010enabling,wang2017visualization}. 

\subsection{Contributions}

Based on the state-of-the-art mentioned above, we list the main contributions of this work in the following.

The first contribution of this work is to develop a new framework for recognizing subterranean junctions using spectral clustering, which is a principled unsupervised learning technique. The proposed method allows us to uncover intrinsic structures from the 2D point cloud extracted from a lidar. Unlike previous techniques, the proposed method applies to various settings without requiring any prior knowledge on the number of junctions and their topology. Furthermore, our method does not require any pre-training and work well when working with small sample sizes. Hence, our introduced framework is suitable for a wide variety of settings with complex geometry. 

In addition to our algorithmic contributions, the second contribution originates from a comprehensive evaluation of the proposed method in simulation environments with complex geometry and data-sets collected from the autonomous flight of the \gls{mav} in real underground mines in Sweden. As we will see, the obtained results demonstrate the effectiveness of our proposed framework on situational awareness of real-world subterranean environments.

The rest of the article is structured as follows. Section \ref{sec:problemstatement} provides a brief overview of detecting junctions in subterranean environments. Section~\ref{sec:methodology} presents our proposed methodology for identifying the number of intersections or junctions. Then, in Section~\ref{sec:results}, we evaluate the proposed method on various simulated and real-world data-sets. Finally, Section~\ref{sec:conclustion} concludes this article by summarizing our findings and offering some future research directions to improve our framework further.

\section{Problem Statement} \label{sec:problemstatement}
\begin{figure}[htbp!]
 \centering
  \includegraphics[width=\linewidth]{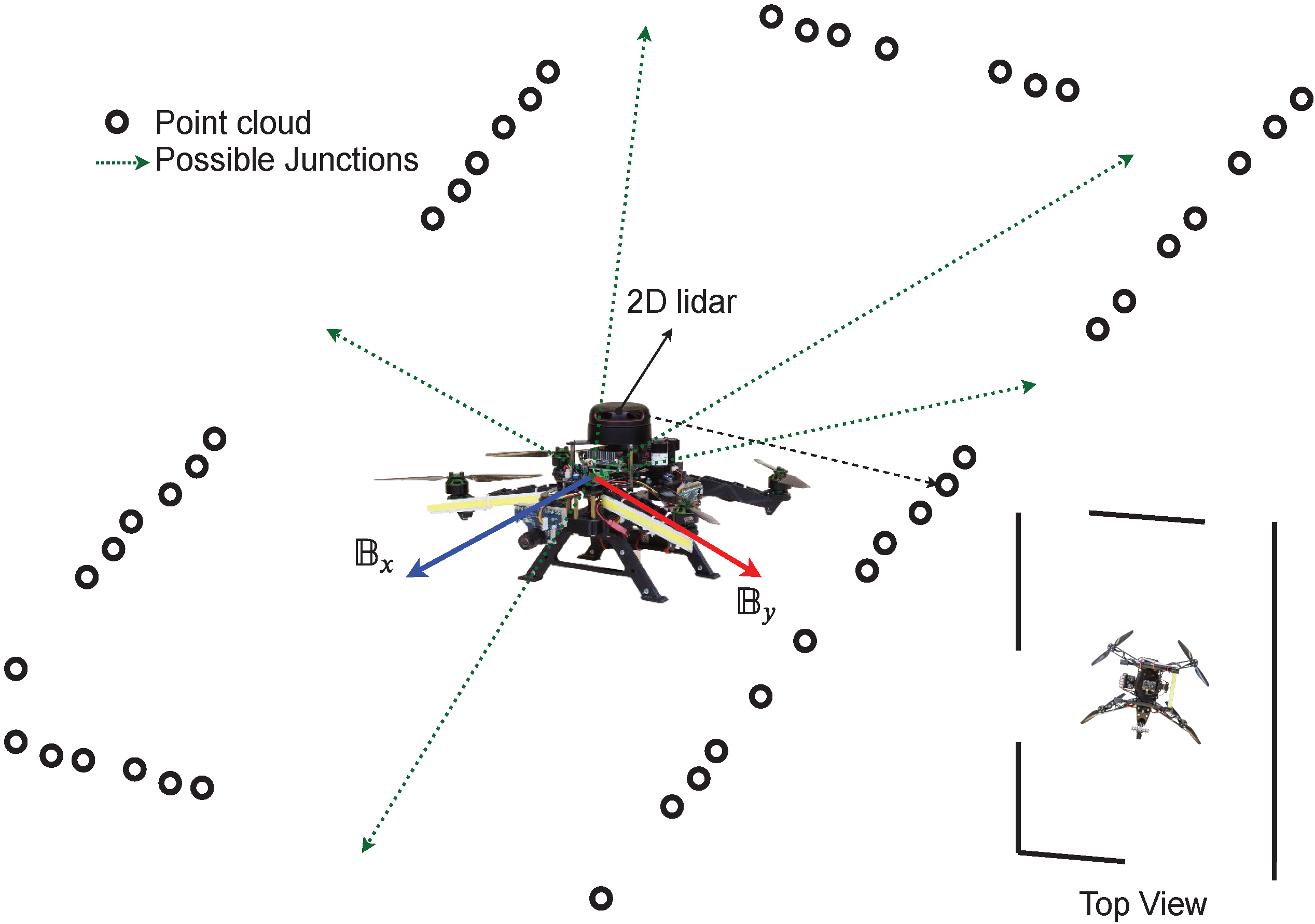} 
   \caption{Top-view concept image of a subterranean environment with multiple branches, while body fixed frame of the \gls{mav} is depicted by $\mathbb{B}$.}
    \label{fig:coordinateframes}    
\end{figure}

This article considers lightweight aerial scouts, which are disposable low-cost robots with the task of fast exploration of unknown subterranean environments for collecting data and provide preliminary feedback for higher level mission planner. These robots should be able to effectively navigate in such challenging environments based on the collected local information. It should be highlighted that most of the works investigating \gls{mav} navigation in subterranean environments consider high end platforms~\cite{papachristos2019autonomous}. In fact, these methods rely on a global map or are evaluated in tunnels without multiple branches~\cite{mansouri2019autonomouscontour}. However, uncertainties and drift over time in localization or map will result in wrong directions, as the concept image depicts in Figure~\ref{fig:coordinateframes}. In this case, the underground tunnel has multiple branches and existing methods have limited success. Thus, the objective of this article is to propose an effective software perspective to detect the number of junctions in man-made tunnels based on the local point cloud information which is obtained from 2D lidars. Thus, our method provides navigation capabilities with resource-constrained solutions. The point cloud is in a body fixed frame of the \gls{mav}, and the \gls{mav} is centered of the point cloud.

\section{Junction Detection} \label{sec:methodology}
In this section, we discuss our proposed methodology for identifying junctions in underground environments by utilizing the acquired 2D point clouds, extracted from a 2D lidar. Let $X\in\mathbb{R}^{n\times 2}$ be a data matrix comprising of $n$ data points. Thus, each row of $X$ corresponds to one data point in $\mathbb{R}^2$ from the given 2D point cloud.
Without any prior information, our goal is to identify the number of junctions, which is equal to the number of walls. In this paper we target the junction recognition with the use of the point cloud generated by only one revolution of the lidar. A wall will be registered as a set of data points with geographical proximity. The goal of this paper is therefore to utilize clustering techniques in order to identify the number of walls and their position. While one may attempt to use simple unsupervised learning techniques, such as $K$-means clustering, a challenge that we face in our problem of interest and it will be further analyzed in Section \ref{sec:problemstatement}, is to handle complex and non-linear structures. Hence, we propose to use spectral clustering techniques that will allow us to capture such complex structures. 

In the following, we explain the three main steps of spectral clustering in detail. Given the data matrix $X$, we first construct a similarity graph. We denote the undirected similarity graph by $G=(V,E,W)$, where we have a set of $n$ nodes $V$ that are interconnected by a set of edges $E$. The matrix $W\in\mathbb{R}^{n\times n}$ is the adjacency matrix of this graph such that $W(i,j)\geq 0$ and $W$ is symmetric. A common approach to construct the similarity graph is to use the \gls{rbf} as a measure of similarity. For some user-defined $\sigma>0$, we compute the pairwise similarity between $x_i$ and $x_j$ (recall that $x_i$ refers to the $i$-th row of $X$):
\begin{align}
W(i,j)=\exp(-\sigma \|x_i-x_j\|_2^2),\; \forall i,j\in\{1,\ldots,n\}.
\end{align}

The next task is to compute a spectral embedding of the original data points $x_1,\ldots,x_n$. Using the adjacency matrix $W$, we form the normalized Laplacian matrix $L\in\mathbb{R}^{n\times n}$ in the following form:
\begin{align}
    L=I - D^{-1/2}W D^{-1/2},
\end{align}
where $I$ is the identity matrix and $D\in\mathbb{R}^{n\times n}$ is a diagonal degree matrix, i.e., $D(i,i)=\sum_{j=1}^{n}W(i,j)$. Hence, the matrix $D^{-1/2}$ can be efficiently formed by computing the inverse square root of the entries of $D$ on the main diagonal.  

The eigenvalue decomposition of $L$ provides valuable insights about the structure of the similarity graph $G$. First, it is worth pointing out that every eigenvalue of the Laplacian matrix is non-negative, i.e., the matrix $L$ is \gls{psd}. Thus, we can represent all $n$ eigenvalues of $L$ in non-decreasing order as follows:
\begin{align}
0=\lambda_1\leq \lambda_2 \leq \ldots \leq \lambda_n.
\end{align}
Furthermore, the number of connected components in the graph $G$ is equal to the multiplicity of the $0$ eigenvalue~\cite{marsden2013eigenvalues}. Hence, spectral clustering does not require any prior knowledge of the number of components. 

Let $k$ be the estimated number of connected components using the above eigenvalue decomposition of the Laplacian. The subsequent task is to partition the original data points $x_1,\ldots,x_n$ into $k$ clusters. To this end, we compute the eigenvectors associated with the $k$ smallest eigenvalues of $L$. Next, we form a new matrix $U\in\mathbb{R}^{n\times k}$ by concatenating these $k$ eigenvectors column-wise. Hence, we can view the $i$-th row of $U$, i.e., $u_i$, as the spectral embedding of the original data point $x_i$. We can thus look for the connected data points in the transformed $k$-dimensional data-set $u_1,\ldots,u_n$ (instead of the original feature space $\mathbb{R}^2$). As mentioned before, this step is essential for many complex data-sets as the original data points may not be linearly separable~\cite{langone2017fast}.

The last step is to perform the \textit{K-means} clustering algorithm on the non-linearly transformed data points $u_1,\ldots,u_n$. The corresponding optimization problem has the following form:  
\begin{align}
   \min_{\mathcal{C}}f(\mathcal{C},U)=\sum_{i=1}^n\min_{c\in \mathcal{C}} \|u_i-c\|_2^2,
\end{align}
where $\mathcal{C}$ represents a set of $k$ cluster centroids or representatives. Solving this optimization problem is known to be NP-hard and thus computationally intractable. However, there are approximation algorithms, such as the \textit{K-means++} algorithm, to solve this problem efficiently. As a result, we will find the partitioning of the embedded data points $u_1,\ldots,u_n$ into $k$ clusters. Note that this assignment can be easily translated to the original data points $x_1,\ldots,x_n$. The proposed algorithm is summarized in Algorithm~\ref{alg:alg1}. 

\begin{algorithm}
  \KwData{A set of $n$ data points $x_1,\ldots,x_n$ in $\mathbb{R}^2$}
  \Parameter{$\sigma>0$}
  \KwResult{Number of junctions and clustering $X$}
  Form the adjacency matrix $W\in\mathbb{R}^{n\times n}$ using a non-linear similarity function, such that $W_{ij}=\exp(-\sigma \|x_i-x_j\|_2^2)$\;
  Form the normalized Laplacian matrix $ L=I - D^{-1/2}W D^{-1/2}$\;
  Compute the eigenvalue decomposition of $L$ and sort the eigenvalues in non-decreasing order\;
  Find the number of zero eigenvalues $k$\;
  Perform \textit{K-means++} clustering on the top $k$ eigenvectors $u_1,\ldots,u_n$ in $\mathbb{R}^k$\; 
  
  \caption{Junction Detection via 2D Point Cloud \label{alg:alg1}}
\end{algorithm}

\section{Results} \label{sec:results}

\subsection{Simulation Evaluations}
\begin{figure*}[htbp!]
    \centering
       \resizebox{\linewidth}{!}{%
\input{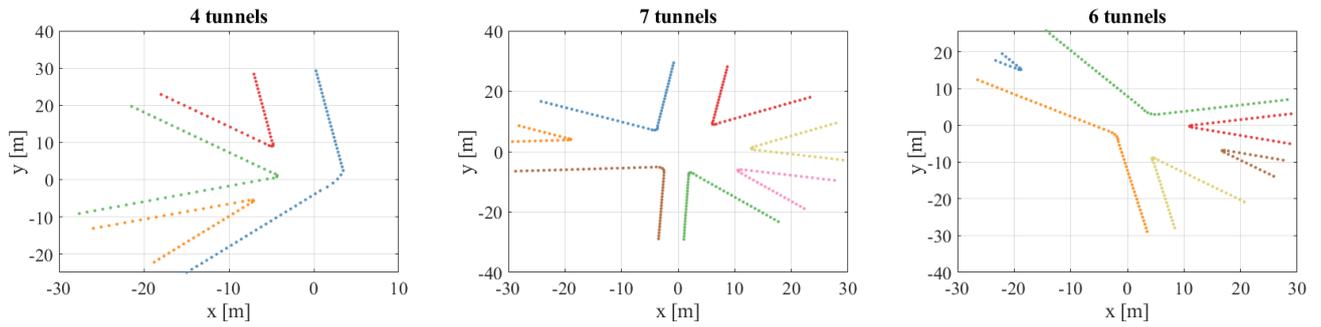}
}
    \caption{Simulation results for identifying the number of junctions in different scenarios using spectral clustering. Each wall is indicated with a different color and the number of junctions is written above each figure (best viewed in color).}
    \label{fig:simulationresults}
\end{figure*}

It should be highlighted that access to subterranean environments is typically limited. Thus, initially for evaluating the proposed method with complex geometry, random range measurements from 2D lidars are generated in the MATLAB~\cite{MATLAB2018}. The 2D lidar is considered with maximum range of $\unit[15]{m}$ and with 360 measurements for each revolution. The obtained point cloud from 2D lidar measurements are fed to the proposed method, the number of junctions are indicated, and the walls are clustered and color coded. All simulations have been performed on a single core on a computer with an Intel Core i7-6600U CPU, 2.6GHz and 8GB RAM and $\sigma = 1.5$, while the average computation time for all scenarios is $\unit[0.2]{s}$. Figure~\ref{fig:simulationresults} depicts the obtained results, while the number of junctions are written above the figures, and the \gls{mav} is considered at the center of the point cloud. The proposed method successfully detects all junctions and walls when considering various types of complex geometry for the simulated 2D point clouds. 


\subsection{Experimental Data-sets}
The data-sets have been collected from autonomous navigation of the \gls{mav} in two different underground mines, while the platform is equipped with a 2D lidar~\cite{mansouri2019visioncnn,mansouri2019autonomouscontour}. The first location is an underground tunnel with a few branches, and the second one is an underground mine with multiple branches. Both environments are located in Sweden. The width and height of the first and second areas are approximately $\unit[3.5]{m} \times \unit[3]{m}$ and $\unit[6]{m} \times \unit[4]{m}$ respectively. 

\begin{figure*}[htbp!]
    \centering
   \resizebox{\linewidth}{!}{%
\input{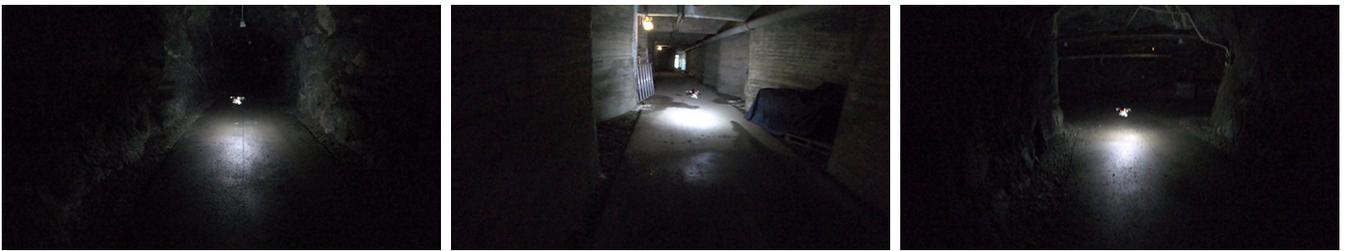}
}
    \caption{The images of the \gls{mav} navigation in three different environments of an underground tunnel in Sweden.}
    \label{fig:luleaexpdata}
\end{figure*}

\begin{figure*}[htbp!]
    \centering
   \resizebox{\linewidth}{!}{%
\input{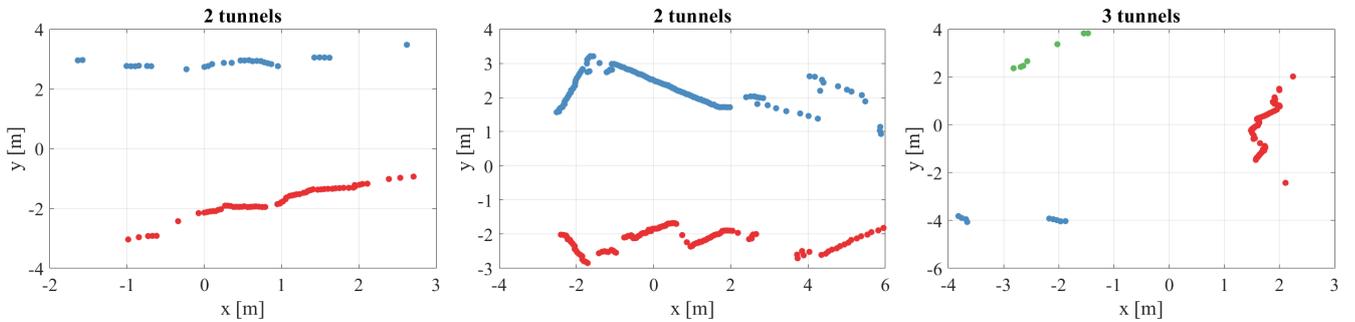}
}
    \caption{The detected walls and junctions from a 2D point cloud utilizing the proposed framework, with experimental data-set from autonomous \gls{mav} navigation in an underground tunnel, while each wall is indicated with a different color.}
    \label{fig:luleaexpdataresults}
\end{figure*}

Figure~\ref{fig:luleaexpdata} depicts the irregular geometry of the first environment, while three specific areas are chosen to evaluate the proposed method, the first and second areas are tunnels with unstructured walls, and the third area has three junctions. The point cloud extracted from 2D lidar is fed to our proposed method and Figure~\ref{fig:luleaexpdataresults} shows the obtained results. The method detects the number of junctions correctly, even though the point cloud extracted from experimental data-set is sparse compare to simulation results. Additionally, it should be highlighted that the tunnel without any branch considers two junctions as the platform has two options to navigate. In the future work, the obtained junctions should be matched to avoid counting same one multiple times and the proposed method should be evaluated in the closed loop with navigation and mission planner modules. 

\begin{figure*}[htbp!]
    \centering
   \resizebox{\linewidth}{!}{%
\input{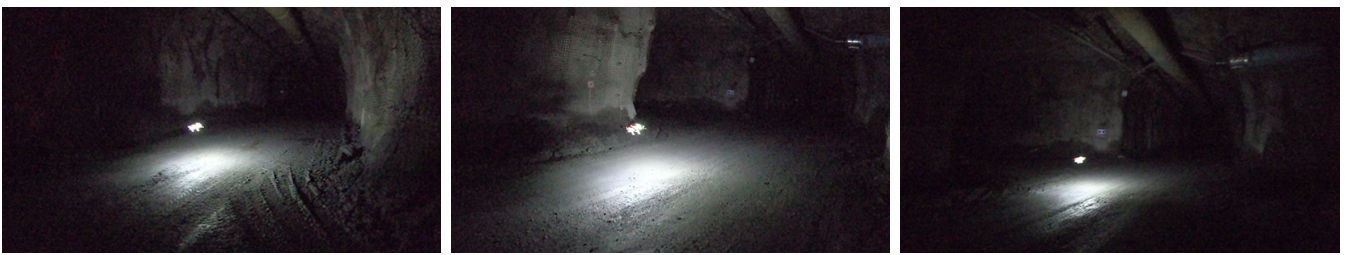}
}
    \caption{The images of the \gls{mav} navigation in three different environments of an underground mine in Sweden.}
    \label{fig:lkabexpdata}
\end{figure*}

\begin{figure*}[htbp!]
    \centering
   \resizebox{\linewidth}{!}{%
\input{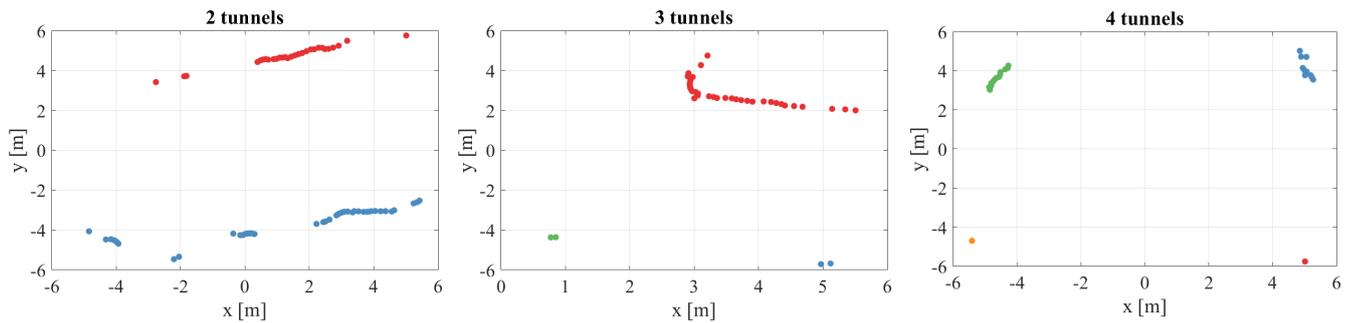}
}
    \caption{The detected walls and junctions from 2D point cloud using the proposed method, with experimental data-set from autonomous \gls{mav} navigation in an underground mine, while each wall is indicated with a different color.}
    \label{fig:lkaexpdataresults}
\end{figure*}

In the second case, the \gls{mav} autonomously navigates in an underground mine, however the width of the area is larger than the first case and the walls have wet surfaces. The areas with three and four junctions are chosen for evaluating the proposed method as depicted in Figure~\ref{fig:lkabexpdata}. Conditions such as the large width and wet wall surfaces result to in a sparse point cloud from 2D lidar range measurements. As it can be seen in Figure~\ref{fig:lkaexpdataresults}, our proposed method detects the number of junctions and walls correctly. Future work should investigate the development of robust methodologies, as dust can be misclassified as junctions. However, in our case, the areas we studied have negligible amount of dust.

\section{Conclusion and Future Work} \label{sec:conclustion}
This article proposed a novel method for the problem of junction detection in subterranean environments. The proposed method uses a non-linear pairwise similarity function to form a graph and find the number of connected components. The proposed method is computationally efficient and detects the number of junctions and walls with average computation time of $\unit[0.2]{s}$, while the detected walls can be used for wall following methods and number of junctions for local or high level planners. The method is successfully evaluated on multiple random generated 2D point clouds and extracted point clouds from autonomous navigation of the \gls{mav} in underground mines. Future works include matching the detected junctions for loop closure and to avoid counting same junctions multiple time, and evaluation of the proposed method in closed loop of navigation and planning components. The proposed unsupervised learning approach in this article will be also extended in the future work to handle 3D point clouds.
\newpage
 \bibliographystyle{IEEEtran}
%
\bibliography{mybib}

\begin{thebibliography}{10}
\providecommand{\url}[1]{#1}
\csname url@samestyle\endcsname
\providecommand{\newblock}{\relax}
\providecommand{\bibinfo}[2]{#2}
\providecommand{\BIBentrySTDinterwordspacing}{\spaceskip=0pt\relax}
\providecommand{\BIBentryALTinterwordstretchfactor}{4}
\providecommand{\BIBentryALTinterwordspacing}{\spaceskip=\fontdimen2\font plus
\BIBentryALTinterwordstretchfactor\fontdimen3\font minus
  \fontdimen4\font\relax}
\providecommand{\BIBforeignlanguage}[2]{{%
\expandafter\ifx\csname l@#1\endcsname\relax
\typeout{** WARNING: IEEEtran.bst: No hyphenation pattern has been}%
\typeout{** loaded for the language `#1'. Using the pattern for}%
\typeout{** the default language instead.}%
\else
\language=\csname l@#1\endcsname
\fi
#2}}
\providecommand{\BIBdecl}{\relax}
\BIBdecl

\bibitem{mansouri2019visioncnn}
S.~S. Mansouri, P.~Karvelis, C.~Kanellakis, D.~Kominiak, and G.~Nikolakopoulos,
  ``Vision-based mav navigation in underground mine using convolutional neural
  network,'' in \emph{IECON 2019-45th Annual Conference of the IEEE Industrial
  Electronics Society}, vol.~1.\hskip 1em plus 0.5em minus 0.4em\relax IEEE,
  2019, pp. 750--755.

\bibitem{tomic2012toward}
T.~Tomic, K.~Schmid, P.~Lutz, A.~Domel, M.~Kassecker, E.~Mair, I.~L. Grixa,
  F.~Ruess, M.~Suppa, and D.~Burschka, ``Toward a fully autonomous uav:
  Research platform for indoor and outdoor urban search and rescue,''
  \emph{IEEE robotics \& automation magazine}, vol.~19, no.~3, pp. 46--56,
  2012.

\bibitem{mansouri2018cooperative}
S.~S. Mansouri, C.~Kanellakis, E.~Fresk, D.~Kominiak, and G.~Nikolakopoulos,
  ``Cooperative coverage path planning for visual inspection,'' \emph{Control
  Engineering Practice}, vol.~74, pp. 118--131, 2018.

\bibitem{von2007tutorial}
U.~Von~Luxburg, ``A tutorial on spectral clustering,'' \emph{Statistics and
  computing}, vol.~17, no.~4, pp. 395--416, 2007.

\bibitem{pourkamali2019improved}
F.~Pourkamali-Anaraki and S.~Becker, ``Improved fixed-rank {N}ystr{\"o}m
  approximation via {QR} decomposition: Practical and theoretical aspects,''
  \emph{Neurocomputing}, vol. 363, pp. 261--272, 2019.

\bibitem{tremblay2020approximating}
N.~Tremblay and A.~Loukas, ``Approximating spectral clustering via sampling: a
  review,'' in \emph{Sampling Techniques for Supervised or Unsupervised Tasks},
  2020, pp. 129--183.

\bibitem{pourkamali2017preconditioned}
F.~Pourkamali-Anaraki and S.~Becker, ``Preconditioned data sparsification for
  big data with applications to {PCA} and {K}-means,'' \emph{IEEE Transactions
  on Information Theory}, vol.~63, no.~5, pp. 2954--2974, 2017.

\bibitem{Kanellakis2017}
C.~Kanellakis and G.~Nikolakopoulos, ``Survey on computer vision for uavs:
  Current developments and trends,'' \emph{Journal of Intelligent {\&} Robotic
  Systems}, pp. 1--28, 2017.

\bibitem{barfoot2016into}
T.~D. Barfoot, C.~McManus, S.~Anderson, H.~Dong, E.~Beerepoot, C.~H. Tong,
  P.~Furgale, J.~D. Gammell, and J.~Enright, ``Into darkness: Visual navigation
  based on a lidar-intensity-image pipeline,'' in \emph{Robotics
  Research}.\hskip 1em plus 0.5em minus 0.4em\relax Springer, 2016, pp.
  487--504.

\bibitem{chaudhuri2012semi}
D.~Chaudhuri, N.~Kushwaha, and A.~Samal, ``Semi-automated road detection from
  high resolution satellite images by directional morphological enhancement and
  segmentation techniques,'' \emph{IEEE journal of selected topics in applied
  earth observations and remote sensing}, vol.~5, no.~5, pp. 1538--1544, 2012.

\bibitem{chen2019higher}
Z.~Chen, C.~Liu, and H.~Wu, ``A higher-order tensor voting-based approach for
  road junction detection and delineation from airborne lidar data,''
  \emph{ISPRS journal of photogrammetry and remote sensing}, vol. 150, pp.
  91--114, 2019.

\bibitem{negri2006junction}
M.~Negri, P.~Gamba, G.~Lisini, and F.~Tupin, ``Junction-aware extraction and
  regularization of urban road networks in high-resolution sar images,''
  \emph{IEEE Transactions on Geoscience and Remote Sensing}, vol.~44, no.~10,
  pp. 2962--2971, 2006.

\bibitem{habermann2016road}
D.~Habermann, C.~E. Vido, F.~S. Os{\'o}rio, and F.~Ramos, ``Road junction
  detection from 3d point clouds,'' in \emph{2016 International Joint
  Conference on Neural Networks (IJCNN)}.\hskip 1em plus 0.5em minus
  0.4em\relax IEEE, 2016, pp. 4934--4940.

\bibitem{mueller2011gis}
A.~Mueller, M.~Himmelsbach, T.~Luettel, F.~v. Hundelshausen, and H.-J.
  Wuensche, ``Gis-based topological robot localization through lidar crossroad
  detection,'' in \emph{2011 14th International IEEE Conference on Intelligent
  Transportation Systems (ITSC)}.\hskip 1em plus 0.5em minus 0.4em\relax IEEE,
  2011, pp. 2001--2008.

\bibitem{kumaar2018juncnet}
S.~Kumaar, S.~Mannar, S.~Omkar \emph{et~al.}, ``Juncnet: A deep neural network
  for road junction disambiguation for autonomous vehicles,'' \emph{arXiv
  preprint arXiv:1809.01011}, 2018.

\bibitem{bhatt2017have}
D.~Bhatt, D.~Sodhi, A.~Pal, V.~Balasubramanian, and M.~Krishna, ``Have i
  reached the intersection: A deep learning-based approach for intersection
  detection from monocular cameras,'' in \emph{2017 IEEE/RSJ International
  Conference on Intelligent Robots and Systems (IROS)}.\hskip 1em plus 0.5em
  minus 0.4em\relax IEEE, 2017, pp. 4495--4500.

\bibitem{kumar2018towards}
A.~Kumar, G.~Gupta, A.~Sharma, and K.~M. Krishna, ``Towards view-invariant
  intersection recognition from videos using deep network ensembles,'' in
  \emph{2018 IEEE/RSJ International Conference on Intelligent Robots and
  Systems (IROS)}.\hskip 1em plus 0.5em minus 0.4em\relax IEEE, 2018, pp.
  1053--1060.

\bibitem{yih2011learning}
W.-t. Yih, K.~Toutanova, J.~C. Platt, and C.~Meek, ``Learning discriminative
  projections for text similarity measures,'' in \emph{Proceedings of the
  fifteenth conference on computational natural language learning}.\hskip 1em
  plus 0.5em minus 0.4em\relax Association for Computational Linguistics, 2011,
  pp. 247--256.

\bibitem{mansouri2019visualjunction}
S.~S. Mansouri, P.~Karvelis, C.~Kanellakis, A.~Koval, and G.~Nikolakopoulos,
  ``Visual subterranean junction recognition for mavs based on convolutional
  neural networks,'' in \emph{IECON 2019-45th Annual Conference of the IEEE
  Industrial Electronics Society}, vol.~1.\hskip 1em plus 0.5em minus
  0.4em\relax IEEE, 2019, pp. 192--197.

\bibitem{alexnet}
\BIBentryALTinterwordspacing
A.~Krizhevsky, I.~Sutskever, and G.~E. Hinton, ``Imagenet classification with
  deep convolutional neural networks,'' pp. 1097--1105, 2012. [Online].
  Available:
  \url{http://papers.nips.cc/paper/4824-imagenet-classification-with-deep-convolutional-neural-networks.pdf}
\BIBentrySTDinterwordspacing

\bibitem{chen2017revisiting}
P.~Chen and L.~Wu, ``Revisiting spectral graph clustering with generative
  community models,'' in \emph{2017 IEEE International Conference on Data
  Mining (ICDM)}, 2017, pp. 51--60.

\bibitem{ng2002spectral}
A.~Ng, M.~Jordan, and Y.~Weiss, ``On spectral clustering: Analysis and an
  algorithm,'' in \emph{Advances in neural information processing systems},
  2002, pp. 849--856.

\bibitem{tung2010enabling}
F.~Tung, A.~Wong, and D.~Clausi, ``Enabling scalable spectral clustering for
  image segmentation,'' \emph{Pattern Recognition}, vol.~43, no.~12, pp.
  4069--4076, 2010.

\bibitem{wang2017visualization}
B.~Wang, J.~Zhu, E.~Pierson, D.~Ramazzotti, and S.~Batzoglou, ``Visualization
  and analysis of single-cell rna-seq data by kernel-based similarity
  learning,'' \emph{Nature methods}, vol.~14, no.~4, p. 414, 2017.

\bibitem{papachristos2019autonomous}
C.~Papachristos, S.~Khattak, F.~Mascarich, and K.~Alexis, ``Autonomous
  navigation and mapping in underground mines using aerial robots,'' in
  \emph{2019 IEEE Aerospace Conference}.\hskip 1em plus 0.5em minus 0.4em\relax
  IEEE, 2019, pp. 1--8.

\bibitem{mansouri2019autonomouscontour}
S.~S. Mansouri, M.~Casta{\~n}o, C.~Kanellakis, and G.~Nikolakopoulos,
  ``Autonomous mav navigation in underground mines using darkness contours
  detection,'' in \emph{International Conference on Computer Vision
  Systems}.\hskip 1em plus 0.5em minus 0.4em\relax Springer, 2019, pp.
  164--174.

\bibitem{marsden2013eigenvalues}
A.~Marsden, ``Eigenvalues of the laplacian and their relationship to the
  connectedness of a graph,'' \emph{University of Chicago, REU}, 2013.

\bibitem{langone2017fast}
R.~Langone and J.~Suykens, ``Fast kernel spectral clustering,''
  \emph{Neurocomputing}, vol. 268, pp. 27--33, 2017.

\bibitem{MATLAB2018}
MATLAB, \emph{version 9.6. (R2019a)}.\hskip 1em plus 0.5em minus 0.4em\relax
  Natick, Massachusetts: The MathWorks Inc., 2019.

\end{thebibliography}
\end{document}